\documentclass[10pt,twocolumn,letterpaper]{article}

\usepackage{iccv}
\usepackage{times}
\usepackage{epsfig}
\usepackage{graphicx}
\usepackage{amsmath}
\usepackage{amssymb}
\usepackage{multirow}
\usepackage{adjustbox}
\usepackage[normalem]{ulem}

\usepackage[breaklinks=true,bookmarks=false]{hyperref}

 \iccvfinalcopy 

\def\iccvPaperID{34} 

\ificcvfinal\fi
\begin{document}
\title{Attack Agnostic Statistical Method for Adversarial Detection}
\author{Sambuddha Saha \qquad Aashish Kumar \qquad Pratyush Sahay \qquad George Jose\\ 
Srinivas Kruthiventi \qquad Harikrishna Muralidhara\\
Harman International India Pvt. Ltd., Bangalore\\
\tt\small$\{$sambuddha.saha, aashish.kumar, pratyush.sahay, george.jose, \\\tt\small srinivas.sai, harikrishna.muralidhara$\}$@harman.com}
\maketitle
\thispagestyle{empty}

\begin{abstract}
   Deep Learning based AI systems have shown great promise in various domains such as vision, audio, autonomous systems (vehicles, drones), etc. Recent research on neural networks has shown the susceptibility of deep networks to adversarial attacks - a technique of adding small perturbations to the inputs which can fool a deep network into misclassifying them. Developing defenses against such adversarial attacks is an active research area, with some approaches proposing robust models that are immune to such adversaries, while other techniques attempt to detect such adversarial inputs. In this paper, we present a novel statistical approach for adversarial detection in image classification. Our approach is based on constructing a per-class feature distribution and detecting adversaries based on comparison of features of a test image with the feature distribution of its class. For this purpose, we make use of various statistical distances such as ED (Energy Distance), MMD (Maximum Mean Discrepancy) for adversarial detection, and analyze the performance of each metric. We experimentally show that our approach achieves good adversarial detection performance on MNIST and CIFAR-10 datasets irrespective of the attack method, sample size and the degree of adversarial perturbation.
\end{abstract}

\section{Introduction}

Deep Learning has been instrumental in the past few years in various domains such as computer vision~\cite{krizhevsky2012imagenet}, audio processing~\cite{hinton2012deep}, natural language processing~\cite{collobert2008unified}~\cite{dahl2012context} and autonomous vehicles~\cite{bojarski2016end}. However, it has recently been shown that these deep networks can be fooled by adding subtle perturbations to the input resulting in misclassification. These perturbed inputs which can still be classified correctly by humans, are known as adversaries~\cite{ExplainingGoodfellow}~\cite{moosavi2016deepfool}.

Two types of approaches are proposed to handle these adversarial attacks. The first approach makes a model robust by training with adversarial examples~\cite{kereliuk2015deep}~\cite{papernot2016limitations}. It applies random perturbations to activations or weights, or it performs feature denoising or by defensive distillation~\cite{papernot2016distillation} to make a model robust to adversaries.

 Other defence approaches based on adversarial detection either use auxiliary networks~\cite{metzen2017detecting} or modify the model architecture and add a detection module and train on adversaries to detect them~\cite{metzen2017detecting}. These approaches are often model centric and are not robust enough for all kinds of attacks. Any network based approach is costly as it involves re-training and customizing the defences for different attacks is also a costly operation. 
 
 Earlier Grosse et. al.~\cite{grosse2017statistical} proposed a statistical based approach which detect adversaries based on the assumption that the original images and the adversarial images belong to two different distributions. They use raw vectorized original images from train set to create a reference distribution and that form the test set to create a test distribution. They create another test distribution from the adversarial images. They compare the two test distributions against the reference distribution using MMD and ED to calculate distances and perform a two sample kernel test to detect if the test distribution belongs to the reference distribution or not. One disadvantage of their experiments is that since they use raw images, the dimensionality is high, so they require samples of higher sizes to approximate the distribution and achieve higher detection confidence (50-100 for the entire dataset). They have also reported per class detection confidence also and they need samples of lower sample size than that for the whole dataset but still it is as high as 20-50 samples.

In this paper, we propose a novel statistical based adversarial detection approach which is agnostic to attacks. Our hypothesis is that the distribution of activations (output of any intermediate layer) of the original data for a particular class is different from that of the adversarial data misclassified into that class. We make use of various statistical metrics to estimate the distance between distributions of the original and the adversarial activations. Adversarial samples will have larger statistical distances from the original distribution and hence can be detected. The proposed approach is attack agnostic (does not vary with the type or degree of attack) and sample efficient (sets with less sample size achieve good detection performance).

\section{Background}
A neural network takes an input $x$ and gives an output, $y$. The outputs are termed as softmax probabilities where $y_i$ denotes the probability of the input $x$ belonging to the class $i$. The softmax probabilities sum up to 1 and lie in the range of 0 to 1. The output label for a particular input, $l(x)$ is assigned by the model as $l(x)=argmax_i(y_i)$  $\forall i \in C$ where $C$ is the total number of classes. The correct label for the class is denoted as $l^* (x)$. The input to the second last layer of the model is termed as \textit{pre-logits} and that to the last softmax layer of the model is termed as \textit{logits}.

Adversarial generation involves perturbing an input $x$ by a small amount to  $x^\prime$ such that the output label of the perturbed input is not same as the output label of the original input, i.e. $l^*(x)\neq l^*(x^\prime)$ where $abs(x - x^\prime) < \epsilon$ where $\epsilon$ is the amount of perturbation and $abs$ represents absolute difference. In the next section we will discuss the various adversarial attacks used in our work.
\subsection{Adversarial Attacks}

\textbf{Fast Gradient Sign Method (FGSM)}: Goodfellow et. al. (2015) ~\cite{ExplainingGoodfellow} proposed this attack where the perturbation $\Delta x$  is based on the gradient of the loss function with respect to the input such that the loss function of the network $C(x,y)$ is maximized. The perturbation is obtained by 
\begin{equation}
\Delta x = \epsilon.sign(\nabla _x C(x,y)) \label{eq:5}
\end{equation}
where $\epsilon$ is the $L_\infty$ norm bound. It is chosen to be small so that $\Delta x$ is undetectable. The $sign$ refers to the direction in which the input feature has to be changed.

\textbf{Carlini-Wagner (CW-$l2$)}: Carlini Wagner et. al. ~\cite{carlini2017towards} proposed an attack using an optimization framework that perturbs the input by inducing very small changes at each iteration to maximize a predefined loss. It generates attack for three different loss metrices, $L_0$, $L_2$ and $L_\infty$. We have used Carlini Wagner $L_2$ attack in this paper.

\textbf{Madry et. al. Attack}: Madry et. al. ~\cite{madry2017towards} proposed a robust optimization based attack to generate adversaries with varying degrees of perturbation, $\epsilon$. They came up with stronger attacks than FGSM using PGD (Projected Gradient Descent).

In the next section, we give a brief description of the various statistical metrics used in this paper.

\subsection{Statistical Metrics}
\textbf{Maximum Mean Discrepancy (MMD)}: Gretton et. al. ~\cite{gretton2012kernel} introduced a kernel based test to compute the distance between probability distributions of two sample sets. The kernel for probability distribution function is chosen such that the difference of the means of the two distributions is maximum.
\begin{equation}
\begin{aligned}
MMD_b[F,X_1,X_2]=sup_{f\in F}(\frac{1}{n}\Sigma_{i=1}^n f(x_{1i})\\-\frac{1}{m}\Sigma_{i=1}^m f(x_{2i})) \label{eq:6}
\end{aligned}
\end{equation}
$X_1$ and $X_2$ refer to the two sample sets and $f$ is the kernel function chosen from $F$ where $F$ represents the super-set of all kernel functions possible. $f$ is chosen to be the kernel which maximizes the difference between the means of the two probability distributions. $x_{1i}$ and $x_{2i}$ denote the probability values of the samples belonging to $X_1$ and $X_2$ respectively for each class $i$ and $m$ and $n$ denotes the number of samples.

\textbf{Energy Distance (ED)}: Szekely et. al. ~\cite{szekely2013energy} proposed an energy based approach to compute distances between two distributions. Let us assume $F$ and $G$ to be two cumulative distribution functions. $X$, $X^\prime$ and $Y$, $Y^\prime$ are independent vectors chosen from $F$ and $G$ respectively which belong to real numbers set $R^d$. The energy distance between the two distributions $F$ and $G$ is the square root of:
\begin{equation}
\begin{aligned}
D^2(F,G)=2E||X-Y||-E||X-X^\prime||\\-E||Y-Y^\prime|| \label{eq:7} 
\end{aligned}
\end{equation}
where $E$ denotes expectation, $||.||$ denotes the norm. ED calculates the distance between two distributions by considering norm distances between samples of different distributions and that of same distribution.

\section{Methodology}

\begin{figure}[h]
\center{\includegraphics[width=0.47\textwidth]{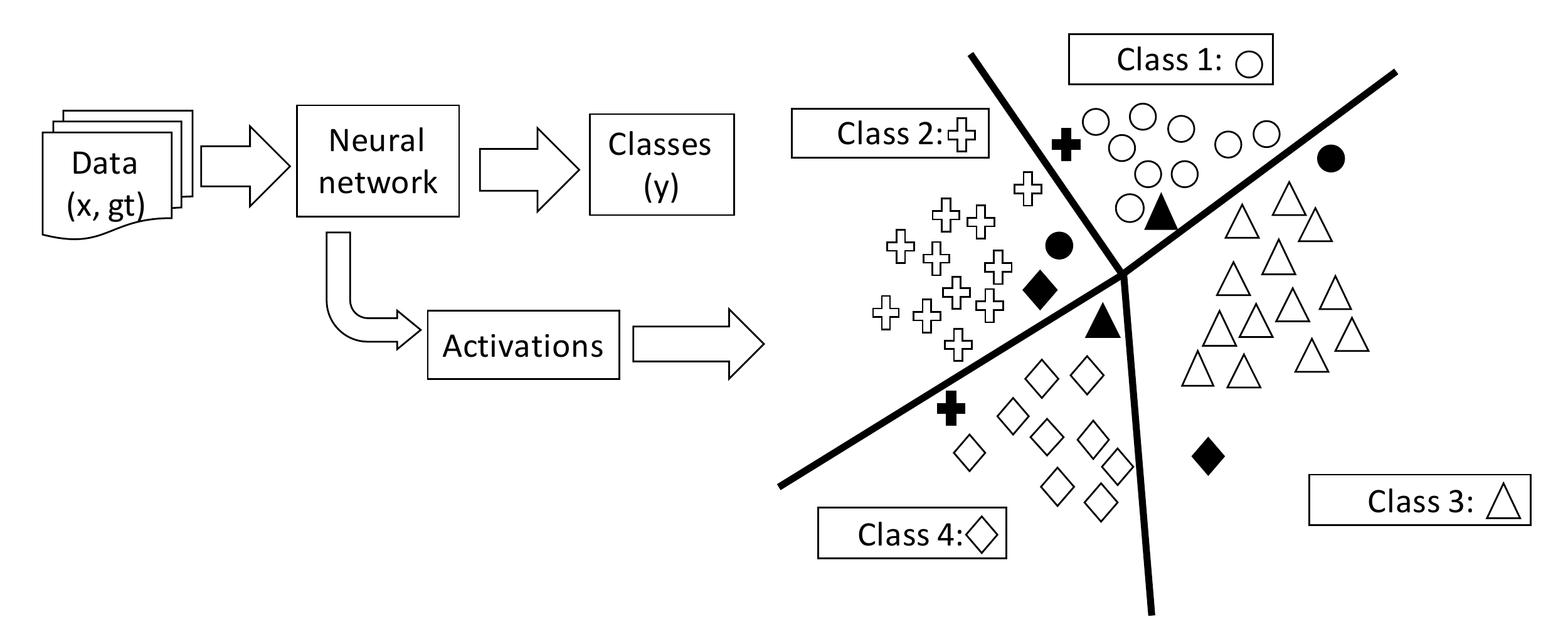}}
   \caption{Illustration of our hypothesis: The activations are extracted from the network for both the original and the adversarial samples. These are shown in a representative plot demarcated by the decision boundaries in the above figure. Adversarial samples which are misclassified are indicated with filled markers. It can be observed that while the activations of original samples belonging to a class cluster together, the adversaries remain as outliers indicating that they do not fit in the distribution.}
   \label{fig:short}
\end{figure}

Our method is based on the hypothesis that the original image activations sample and the adversarial image activations sample belong to two different distributions. We performed a statistical distance based analysis to differentiate between the original and adversarial activations distribution for each class.

Fig.\ref{fig:short} shows a brief overview of the methodology we are following. The model is trained on the data$(x, gt)$ where $x$ is the input image and $gt$ is the ground truth label. We store the output labels and extract the activations (hidden layer activations) from the model. The activations for each class are generally clustered together and each cluster represent different classes as shown in the figure. As observed from the figure, the partition lines are the decision boundaries. When this model is attacked by adversarial samples, the adversarial sample activations lie far away from the original class activations distribution and are misclassified as another class. 

The triangles (refer Fig. \ref{fig:short}) refer to Class 3 original activations distribution where the samples belonging to that are clustered together. The adversarial samples to this class, like the class 4 sample (triangle inside diamond space) or the class 1 sample (triangle inside the circle space) lie further away from the original distribution.

\section{Experiments}
We perform experiments to validate our hypothesis on MNIST (Modified National Institute of Standards and Technology) ~\cite{lecun1998gradient} and CIFAR-10 (Canadian Institute For Advanced Research) ~\cite{krizhevsky2009learning}datasets.

\subsection{Network Setup}

The table below shows the model architecture used for MNIST (refer Table \ref{tab:table_mnist_arch}). We use the default convolution neural network present in the cleverhans repositiory~\cite{papernot2016cleverhans}. 

\begin{table}[h]
\begin{center}
\begin{tabular}{|l|c|c|c|c|}
\hline
ID & Layer Type & Kernel  & \# O/p  & Stride \\
 &  & Size & Channels &  \\
\hline
1 & Conv,Relu & 8 & 32 & 2 \\
\hline
2 &	Conv,Relu &	6 & 64 & 2\\
\hline
3 & Conv, Relu & 3 & 128 & 1 \\
\hline
4 & Conv, Relu & 2	& 128 & 1 \\
\hline
5 & Dense &  & 256 &   \\
\hline
7 & Dense &	 & 128 & 	\\
\hline

\end{tabular}
\end{center}
\caption{Model architecture for MNIST}
\label{tab:table_mnist_arch}
\end{table}

The neural network is trained for 220 epochs with 0.001 learning rate and batch size 128 using Adam optimizer.

We used the same network as above for CIFAR-10 but increased the number of channels in each convolution layers by 4 times. This neural network is trained on the training dataset for 400 epochs, with learning rate 0.001, Adam optimizer and batch size 128. After the training is over, we calculate the accuracy of the model on the test set.

We discard the misclassified samples from the test data after accuracy evaluation as these might lead to false adversaries in adversarial set.

\begin{table*}[h]
\centering
\begin{adjustbox}{width=1\textwidth}
\begin{tabular}{|c|c|c|c|c|c|c|}
\hline
\multirow{2}{*}{Sample Sizes} & \multicolumn{2}{c|}{FGSM} & \multicolumn{2}{c|}{MADRY} & \multicolumn{2}{c|}{CARLINI WAGNER} \\ \cline{2-7} 
 & ED & MMD & ED & MMD & ED & MMD \\ \hline
1 & 98.7 ($\pm$ 0.4) & 97.9 ($\pm$ 0.006) & 99.5 ($\pm$ 0.2) & 98.9 ($\pm$ 0.005) & 100 ($\pm$ 0) & 100 ($\pm$ 0) \\ \hline
5 & 99.6 ($\pm$ 0.04) & 98.4 ($\pm$ 0.002) & 99.7 ($\pm$ 0.02) & 98.97 ($\pm$ 0.0014) & 99.9 ($\pm$ 0) & 99.99 ($\pm$ 0) \\ \hline
10 & 99.7 ($\pm$ 0.03) & 99.1 ($\pm$ 0.001) & 99.8 ($\pm$ 0.01) & 99.4 ($\pm$ 0.0007) & 99.9 ($\pm$ 0) & 99.98 ($\pm$ 0) \\ \hline
20 & 99.8 ($\pm$ 0.01) & 99.4 ($\pm$ 0.0005) & 99.8 ($\pm$ 0) & 99.6 ($\pm$ 0.0004) & 99.9 ($\pm$ 0) & 99.99 ($\pm$ 0) \\ \hline
\end{tabular}
\end{adjustbox}
\caption{AUC scores (\%) for MNIST dataset.}
\label{tab:mnist_res}
\end{table*}

\begin{table*}[h]
\centering
\begin{adjustbox}{width=1\textwidth}
\begin{tabular}{|c|c|c|c|c|c|c|}
\hline
\multirow{2}{*}{Sample Sizes} & \multicolumn{2}{c|}{FGSM} & \multicolumn{2}{c|}{MADRY} & \multicolumn{2}{c|}{CARLINI WAGNER} \\ \cline{2-7} 
 & ED & MMD & ED & MMD & ED & MMD \\ \hline
1 & 75.9 ($\pm$ 2.66) & 71.5 ($\pm$ 0.03) & 88.4 ($\pm$ 2.68) & 87.4 ($\pm$ 0.03) & 94.2 ($\pm$ 2.08) & 92.6 ($\pm$ 0.03) \\ \hline
5 & 83.1 ($\pm$ 0.4) & 76.5 ($\pm$ 0.02) & 91.9 ($\pm$ 0.13) & 88.5 ($\pm$ 0.006) & 94.5 ($\pm$ 0.22) & 92.6 ($\pm$ 0.008) \\ \hline
10 & 84.6 ($\pm$ 0.42) & 83.11 ($\pm$ 0.01) & 92.3 ($\pm$ 0.11) & 90.1 ($\pm$ 0.003) & 94.8 ($\pm$ 0.19) & 94.6 ($\pm$ 0.004) \\ \hline
20 & 87.4 ($\pm$ 0.3) & 87.4 ($\pm$ 0.009) & 92.9 ($\pm$ 0.09) & 91.2 ($\pm$ 0.003) & 95.4 ($\pm$ 0.11) & 95.8 ($\pm$ 0.003) \\ \hline
\end{tabular}
\end{adjustbox}
\caption{AUC scores (\%) for CIFAR-10 dataset.}
\label{tab:cifar_res}
\end{table*}

\subsection{Adversarial Attack Generation and Activations Extraction}
We attack our model using three adversarial attack generation techniques FGSM, Madry and Carlini Wagner. FGSM and Madry attack are generated for 5 varying degrees of perturbations (epsilons), 0.01, 0.05, 0.1, 0.2 and 0.3. We generate adversaries on the correctly classified samples of test data only.

Here we describe the method for activations extraction and distribution generation. We store the \textit{original labels} (ground truth labels for original sample) and extract \textit{logits} and \textit{pre-logits} from the trained model. We store the \textit{adversarial labels} (predicted labels for adversarial sample) and extract \textit{logits} and \textit{pre-logits} from the model. Here we have considered logits as the activations for original and adversarial samples. The original activations are partitioned into a \textit{baseline holdout set} and \textit{rest of the original activations} in another set. The \textit{baseline holdout set} is of fixed size having 100 samples or half the number of total original samples present for that class. The rest represents the \textit{rest of the original activations} set. So we have three activations set  now, \textit{baseline holdout set}, \textit{rest of the original activations} set and the adversarial activations set. The \textit{baseline holdout set} is our \textit{reference set}. The other 2 sets are our \textit{test sets}. Each set is a 2-D matix made up of 1-D activation vectors corresponding to each image. We apply softmax over the activation vector for a sample and perform this for all samples present in all the three sets. A fixed size set of randomly sampled samples is picked from adversarial activations set and \textit{rest of the original activations} set. This is our test sample size which tells us if that number of samples is enough to distinguish between original and adversarial samples.

\subsection{Statistical Distances Computation and AUC Scores Generation}
We calculate statistical metrics for \textit{rest of the original activations} set w.r.t the \textit{reference set} and for the adversarial activations distribution w.r.t the \textit{reference set}. We repeat the above operation 100 times, each time randomly sampling our \textit{test sets} for a particular sample size. We compute AUC score for a particular class for a particular sample size and degree of perturbation. The AUC scores indicate how well the original and adversarial samples can be separated. These scores are tallied for varying sample sizes, degrees of perturbation and different classes.

\section{Results and Discussion}
We present our results on MNIST and CIFAR-10 with three different kinds of attack FGSM, Carlini Wagner and Madry using two statistical distances, MMD and ED. We compute the mean and standard deviation of AUC scores across all the classes and all the epsilons (degrees of perturbation), 0.01, 0.05, 0.1, 0.2, 0.3.
To maintain the brevity of the paper we are showing results corresponding to test sample sizes 1,5,10 and 20.

\subsection{\textbf{MNIST}}
We trained our model on MNIST and it gave a test accuracy of 99.44\%. The AUC scores for MMD and ED for each attack were similar for each sample sizes. Both the statistical distances perform better for strong attacks like Madry and Carlini Wagner than FGSM. The AUC scores increase with increase in test sample size. (see Table \ref{tab:mnist_res}).

\subsection{\textbf{CIFAR-10}}
Our model trained on CIFAR-10 gave a test accuracy of 71.8\% which isn't high enough but surprisingly AUC scores were really good for both the statistical distances for different attacks. We observed similar trends on CIFAR10 as that on MNIST. (see Table \ref{tab:cifar_res})

\subsection{Discussion}
We obtained the following insights by analysing our results 1) The AUC scores obtained using ED and MMD were high and similar for our model across all the three attacks and two datasets. 2) The AUC scores increase proportionally with increase in sample size of the \textit{test set} (works well for test sample of size 1 also) as expected. 3) The AUC scores vary negligibly with change in the degree of attack. 

Hence our model is attack agnostic, which means it doesn't vary with the kind of attack and degree of perturbation. Our model is sample efficient because we experimentally demonstrated that even if the size of our test sample set is one, we are able to achieve good detection.

Since the statistical distances(ED and MMD) perform so well in separating the original and adversarial distributions, it proves our hypothesis that the adversaries don't belong to the same distribution as the natural image distribution and hence can be separated by such statistical distance metrics. The results also prove that the learnt features extracted from the model which are low-dimensional, provides a good approximation of the data. Hence we don't need samples of large sizes to get high detection performance.

\section{Conclusion}
We experimentally demonstrated that the original and adversarial sample do not belong to the same distribution. We also experimentally validated our approach to be attack agnostic and sample efficient. We could expand this work to include more statistical distance metrics and also can extend to use pre-logits. More research will surely contribute to coming up with better statistical models for detecting adversaries.
\clearpage
{\small
\bibliographystyle{ieee}
\bibliography{egbib}

\begin{thebibliography}{10}\itemsep=-1pt

\bibitem{bojarski2016end}
M.~Bojarski, D.~Del~Testa, D.~Dworakowski, B.~Firner, B.~Flepp, P.~Goyal, L.~D.
  Jackel, M.~Monfort, U.~Muller, J.~Zhang, et~al.
\newblock End to end learning for self-driving cars.
\newblock {\em arXiv preprint arXiv:1604.07316}, 2016.

\bibitem{carlini2017towards}
N.~Carlini and D.~Wagner.
\newblock Towards evaluating the robustness of neural networks.
\newblock In {\em Security and Privacy (SP), 2017 IEEE Symposium on}, pages
  39--57. IEEE, 2017.

\bibitem{collobert2008unified}
R.~Collobert and J.~Weston.
\newblock A unified architecture for natural language processing: Deep neural
  networks with multitask learning.
\newblock In {\em Proceedings of the 25th international conference on Machine
  learning}, pages 160--167. ACM, 2008.

\bibitem{dahl2012context}
G.~E. Dahl, D.~Yu, L.~Deng, and A.~Acero.
\newblock Context-dependent pre-trained deep neural networks for
  large-vocabulary speech recognition.
\newblock {\em IEEE Transactions on audio, speech, and language processing},
  20(1):30--42, 2012.

\bibitem{ExplainingGoodfellow}
I.~Goodfellow, J.~Shlens, and C.~Szegedy.
\newblock Explaining and harnessing adversarial examples.
\newblock 12 2014.

\bibitem{gretton2012kernel}
A.~Gretton, K.~M. Borgwardt, M.~J. Rasch, B.~Sch{\"o}lkopf, and A.~Smola.
\newblock A kernel two-sample test.
\newblock {\em Journal of Machine Learning Research}, 13(Mar):723--773, 2012.

\bibitem{grosse2017statistical}
K.~Grosse, P.~Manoharan, N.~Papernot, M.~Backes, and P.~McDaniel.
\newblock On the (statistical) detection of adversarial examples.
\newblock {\em arXiv preprint arXiv:1702.06280}, 2017.

\bibitem{hinton2012deep}
G.~Hinton, L.~Deng, D.~Yu, G.~E. Dahl, A.-r. Mohamed, N.~Jaitly, A.~Senior,
  V.~Vanhoucke, P.~Nguyen, T.~N. Sainath, et~al.
\newblock Deep neural networks for acoustic modeling in speech recognition: The
  shared views of four research groups.
\newblock {\em IEEE Signal processing magazine}, 29(6):82--97, 2012.

\bibitem{kereliuk2015deep}
C.~Kereliuk, B.~L. Sturm, and J.~Larsen.
\newblock Deep learning and music adversaries.
\newblock {\em IEEE Transactions on Multimedia}, 17(11):2059--2071, 2015.

\bibitem{krizhevsky2009learning}
A.~Krizhevsky, G.~Hinton, et~al.
\newblock Learning multiple layers of features from tiny images.
\newblock Technical report, Citeseer, 2009.

\bibitem{krizhevsky2012imagenet}
A.~Krizhevsky, I.~Sutskever, and G.~E. Hinton.
\newblock Imagenet classification with deep convolutional neural networks.
\newblock In {\em Advances in neural information processing systems}, pages
  1097--1105, 2012.

\bibitem{lecun1998gradient}
Y.~LeCun, L.~Bottou, Y.~Bengio, P.~Haffner, et~al.
\newblock Gradient-based learning applied to document recognition.
\newblock {\em Proceedings of the IEEE}, 86(11):2278--2324, 1998.

\bibitem{madry2017towards}
A.~Madry, A.~Makelov, L.~Schmidt, D.~Tsipras, and A.~Vladu.
\newblock Towards deep learning models resistant to adversarial attacks.
\newblock {\em arXiv preprint arXiv:1706.06083}, 2017.

\bibitem{metzen2017detecting}
J.~H. Metzen, T.~Genewein, V.~Fischer, and B.~Bischoff.
\newblock On detecting adversarial perturbations.
\newblock {\em arXiv preprint arXiv:1702.04267}, 2017.

\bibitem{moosavi2016deepfool}
S.-M. Moosavi-Dezfooli, A.~Fawzi, and P.~Frossard.
\newblock Deepfool: a simple and accurate method to fool deep neural networks.
\newblock In {\em Proceedings of the IEEE Conference on Computer Vision and
  Pattern Recognition}, pages 2574--2582, 2016.

\bibitem{papernot2016cleverhans}
N.~Papernot, I.~Goodfellow, R.~Sheatsley, R.~Feinman, and P.~McDaniel.
\newblock cleverhans v2. 0.0: an adversarial machine learning library.
\newblock {\em arXiv preprint arXiv:1610.00768}, 10, 2016.

\bibitem{papernot2016limitations}
N.~Papernot, P.~McDaniel, S.~Jha, M.~Fredrikson, Z.~B. Celik, and A.~Swami.
\newblock The limitations of deep learning in adversarial settings.
\newblock In {\em Security and Privacy (EuroS\&P), 2016 IEEE European Symposium
  on}, pages 372--387. IEEE, 2016.

\bibitem{papernot2016distillation}
N.~Papernot, P.~McDaniel, X.~Wu, S.~Jha, and A.~Swami.
\newblock Distillation as a defense to adversarial perturbations against deep
  neural networks.
\newblock In {\em Security and Privacy (SP), 2016 IEEE Symposium on}, pages
  582--597. IEEE, 2016.

\bibitem{szekely2013energy}
G.~J. Sz{\'e}kely and M.~L. Rizzo.
\newblock Energy statistics: A class of statistics based on distances.
\newblock {\em Journal of statistical planning and inference},
  143(8):1249--1272, 2013.

\end{thebibliography}
}

\end{document}